%% file: main.tex
\documentclass[conference]{IEEEtran}
\IEEEoverridecommandlockouts
\usepackage{cite}
\usepackage{amsmath,amssymb,amsfonts}
\usepackage{algorithmic}
\usepackage{graphicx}
\usepackage{textcomp}
\usepackage{xcolor}
\usepackage{bm}
\usepackage{balance}
\usepackage{xspace}
\usepackage{enumitem}
\usepackage{CJK}
\usepackage{multirow}
\usepackage{bbding} 
\usepackage{booktabs}
\usepackage{hyperref}

\hypersetup{hidelinks} 

\newcommand{\model}{CausalTAD\xspace}
\newcommand{\ie}{\emph{i.e.}\xspace} % and others
\newcommand{\etc}{\emph{etc.}\xspace} % and others
\newcommand{\define}[3]{\vspace{1ex}\noindent{ \textbf{\textsc{Definition {#1}}} (#2): \emph{#3}\vspace{1ex}}
}

\def\BibTeX{{\rm B\kern-.05em{\sc i\kern-.025em b}\kern-.08em
    T\kern-.1667em\lower.7ex\hbox{E}\kern-.125emX}}
\begin{document}

\title{\model: Causal Implicit Generative Model for Debiased Online Trajectory Anomaly Detection}

\author{\IEEEauthorblockN{Wenbin Li$^{1,2}$, $^*$Di Yao$^{1}$, Chang Gong$^{1,2}$, Xiaokai Chu$^1$, Quanliang Jing$^{1}$\\ 
Xiaolei Zhou$^3$, Yuxuan Zhang$^3$, Yunxia Fan$^3$, $^*$Jingping Bi$^1$}
\IEEEauthorblockA{$^1$Institute of Computing Technology, Chinese Academy of Sciences, Beijing, China,\\
$^2$University of Chinese Academy of Sciences, China, \\
$^{3}$DiDi Global Inc.}
\{liwenbin20z,yaodi,gongchang21z,chuxiaokai,jingquanliang,bjp\}@ict.ac.cn, \\
\{zhouxiaolei,tianjinmouthzhangyx,fanyunxia\}@didiglobal.com
\thanks{$^*$Corresponding authors.}
}

\maketitle

\begin{abstract}
Trajectory anomaly detection, aiming to estimate the anomaly risk of trajectories given the Source-Destination (SD) pairs, has become a critical problem for many real-world applications. Existing solutions directly train a generative model for observed trajectories and calculate the conditional generative probability $P(\bm{T}|\bm{C})$ as the anomaly risk, where $\bm{T}$ and $\bm{C}$ represent the trajectory and SD pair respectively. However, we argue that the observed trajectories are confounded by road network preference which is a common cause of both SD distribution and trajectories. Existing methods ignore this issue limiting their generalization ability on out-of-distribution trajectories. In this paper, we define the debiased trajectory anomaly detection problem and propose a causal implicit generative model, namely \model, to solve it. \model adopts \emph{do}-calculus to eliminate the confounding bias of road network preference and estimates $P(\bm{T}|do(\bm{C}))$ as the anomaly criterion.  Extensive experiments show that \model can not only achieve superior performance on trained trajectories but also generally improve the performance of out-of-distribution data, with improvements of $2.1\% \sim 5.7\%$ and $10.6\% \sim 32.7\%$ respectively.
\end{abstract}

% \begin{IEEEkeywords}

% \end{IEEEkeywords}

\input{sections/introduction}
\input{sections/relatedwork}
\input{sections/preliminary}
\input{sections/causalanalysis}
\input{sections/methodology}
\input{sections/experiment}
\input{sections/conclusion}

\section*{Acknowledgements}
This work has been supported by the National Natural Science Foundation of China under Grant No. 62002343. This work is also sponsored by CCF-DiDi GAIA Collaborative Research Funds for Young Scholars.

\balance
\bibliographystyle{IEEEtran}
\bibliography{IEEEabrv,reference}

\end{document}

%% file: sections/introduction.tex
\section{Introduction}
With the rapid growth of ride-hailing services, trajectory data has been collected at an unprecedented speed leading to the flourishing of related research~\cite{zheng2015trajectory,abs-2303-05012,ChenZLYZ22}. To ensure trip safety, trajectory anomaly detection, distinguishing anomalous ongoing detour trajectories under the given source-destination (SD) pairs, is critical yet fundamental for ride-hailing platforms. In practice, detecting anomalies with partial trajectories is more valuable compared with the fully observed manner. The platform would have more time and probability to prevent them from escalating into severe incidents. Thus, online detection is essential for trajectory anomaly detection.

In recent years, various methods have been proposed for online trajectory anomaly detection. Existing works can be roughly divided into two categories, \ie, metric-based methods~\cite{cheng2019stl,zhang2011ibat} and learning-based methods~\cite{gray2018coupled}. Given historical trajectory data, metric-based methods first build an index of SD pair and define a metric to measure the distances between different trajectories. When new trajectories come in, they employ the metric to calculate the trajectory distances between new trajectories and historical trajectories with the same SD pairs. Then, the anomaly trajectories can be filtered by an ad-hoc threshold of distances. The main weakness of these methods is their inability to model the context information, such as traffic congestion, weather conditions, and driver behavior, leading to high false positive rates. To solve this problem, learning-based methods are proposed and achieve superior performance. Taking the context information into account, they train a generation model of trajectory distribution $P(\bm{T})$ and employ the generation probabilities of new trajectories condition on the SD pairs $P(\bm{T}|\bm{C})$ to determine whether the trajectory is abnormal. However, all learning-based methods can only detect the trajectories whose SD pairs occurred in the historical trajectory dataset and perform poorly on unobserved SD pairs, causing the out-of-distribution generalization problem. This problem limits the practical application of these methods, as the training dataset cannot cover all potential SD pairs, and there will always be unseen SD pairs in reality.

\begin{figure}
    \centering
    \includegraphics[height=3.4cm]{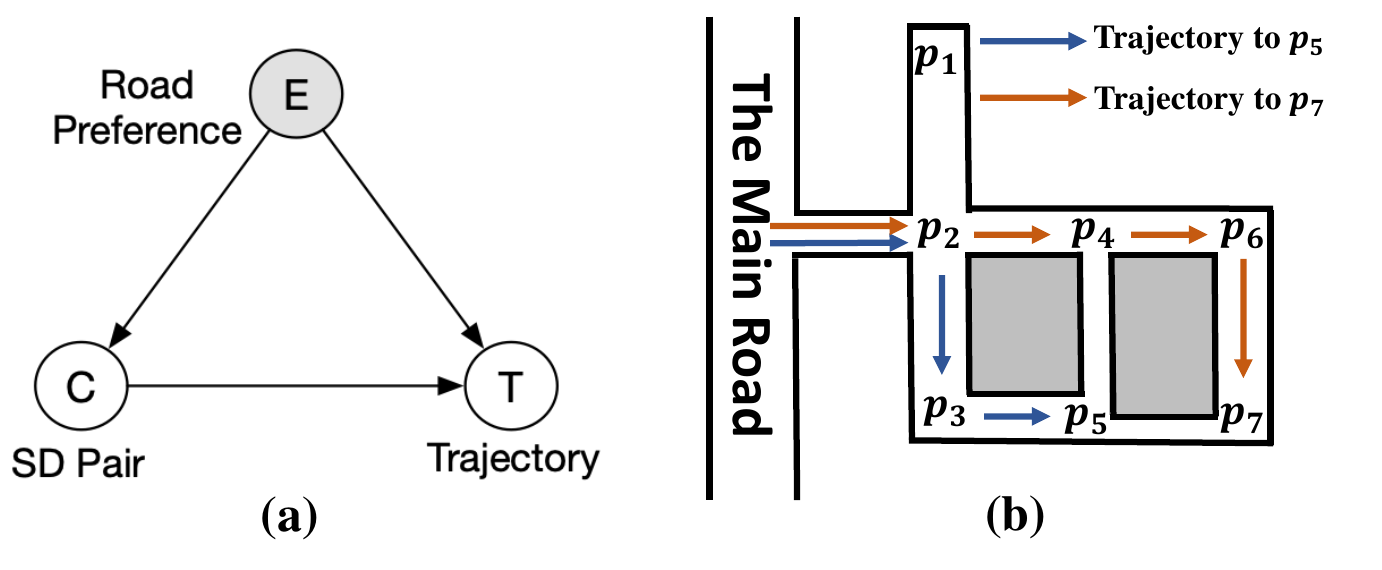}
    \vspace{-2ex}
    \caption{(a) The causal graph of trajectory generation. $\bm{E}$ is the common cause of $\bm{C}$ and $\bm{T}$, which introduces the confounding bias. (b) An example to illustrate the road preference bias.}
    \label{fig:causal_model}
    \vspace{-2ex}
\end{figure}

We attribute the failure of existing methods in generalizing out of distribution to the existence of a confounder, \ie, the road preference, which is a common cause of both observed trajectories and the distribution of SD pairs. Unfortunately, all the generative models ignore the confounding bias and directly learn from the observed trajectories, resulting in additional spurious correlation implicated in the estimation of $P(\bm{T}|\bm{C})$.

To explain the confounding bias, we illustrate the causal graph of the trajectory generation process in Fig.~\ref{fig:causal_model}(a). As shown, $\bm{C}$ and $\bm{T}$ represent the distribution of SD pairs and the observed trajectories respectively. $\bm{E}$ is the road preference reflecting the mixture effects of many factors such as the weather, road level, speed limit and \etc. Road preference and SD pair are the causes of trajectory generation. There also exists a causal edge $\bm{E} \rightarrow \bm{C}$ denoting the SD pairs are affected by road preference. For example, as shown in Fig.~ \ref{fig:causal_model}(b), there is a mall at $p_5$, according to $\bm{E} \rightarrow \bm{C}$, many trajectories in the training dataset are destined for $p_5$. Consider a trajectory coming from the main road and heading for $p_5$, and it has now arrived at $p_2$. According to $\bm{C} \rightarrow \bm{T}$, the driver won't drive into road $p_2-p_1$, since this road is not accessible to the destination $p_5$.
While both road $p_2-p_4$ and road $p_2-p_3$ can lead to the destination $p_5$, the driver prefers $p_2-p_3$ since $p_2-p_4-p_5$ is narrower and more congested than $p_2-p_3-p_5$, which conforms to $\bm{E}\rightarrow \bm{T}$. The conditional probability $P(\bm{C}|\bm{T})$ models the correlation between $\bm{C}$ and $\bm{T}$ caused by both $\bm{C} \rightarrow \bm{T}$ and $\bm{C} \leftarrow \bm{E} \rightarrow \bm{T}$, but it cannot distinguish them. When there are new SD pairs during inference, the correlation caused by $\bm{C} \leftarrow \bm{E} \rightarrow \bm{T}$ may not hold, resulting in the failure of existing methods. 
% For example, the model may mistakenly believe that normal trajectories prefer the road $p_2-p_3$ over $p_2-p_4$ according to the trajectories in the training dataset. 
For example, consider a trajectory heading for $p_7$ in Fig.~\ref{fig:causal_model}(b), due to the very narrow road $p_5-p_7$, the normal route is $p_2-p_4-p_6-p_7$ and drivers prefer $p_2-p_4$ over $p_2-p_3$, which contradicts the road preferences in the training dataset. As a result, for the new destination $p_7$, the model will assign a very low probability $P(\bm{T}|\bm{C})$ to a normal trajectory according to the road $p_2-p_4$, and it finally predicts the normal trajectory as an anomaly. With sufficient data for each SD pair, this confounding bias could be mitigated. Nevertheless, it is infeasible to collect sufficient data for all SD pairs, leading to the poor performances of the existing method on unobserved SD pairs. Therefore, there is an urgent need for debiased online trajectory anomaly detection.

However, eliminating the confounding bias of $\bm{E}$ is not trivial. Compared with the traditional problem, debiased online trajectory anomaly detection introduces three challenges: (1) Hidden confounder. As shown in Fig.~\ref{fig:causal_model}(a), the road preference $\bm{E}$ is an unobserved hidden confounder representing the combination influence of different road-related factors. It introduces the spurious correlations of $\bm{C} \rightarrow \bm{T}$ which is hard to be mitigated by classical debiasing methods \cite{caliendo2008some, mansournia2016inverse}. 
% (2) Road network constrained. Both normal and abnormal trajectories are constrained by the road network. How to integrate the road network constraint into the generative model is challenging for this task. 
(2) Out-of-distribution generalization. Using historical trajectories of limit SD pairs, the debiased anomaly detection aims to estimate the generation probabilities of trajectories of any SD pairs. How to deal with trajectories of unobserved SD pairs is challenging for this task. 
(3) Efficiency. Efficiency is essential for online anomaly detection. Thus, the elimination of confounding bias should be efficient and the debiased anomaly risk of an ongoing trajectory should be updated with the time complexity of $\mathcal{O}(1)$.

In this paper, we proposed \model to address these challenges systematically. Specifically, we adopt \emph{do}-calculus to define $P(\bm{T}|do(\bm{C}))$ as the debiased anomaly criterion that characterizes the causality of trajectory generation. Then \model estimates the anomaly criterion via two key components, \ie Trajectory Generation VAE (TG-VAE), and Road Preference VAE (RP-VAE). The TG-VAE aims to learn the trajectory patterns for each SD pair. It infers the hidden representation according to the SD pair and predicts the trajectory conditioned on the hidden representation in an autoregressive way under the constraint of the road network. The RP-VAE factorizes the debiasing scaling factor of a whole trajectory to each road segment, and approximates it via the probability likelihood of each road segment.
For efficiency, the scaling factors can be calculated and stored in advance during inference to support online anomaly detection.

The contribution of this paper can be summarized as follows:

\begin{itemize}[leftmargin=4mm]
    \item We provide a detailed analysis of the trajectory generation process under the causal perspective. To the best of our knowledge, this is the first work finding that both the distribution of SD pairs and observed trajectories are caused by a hidden confounder, \ie road preference. 
    \item Based on the analysis, we define a new task, namely debiased online trajectory anomaly detection, and propose a novel causal implicit generative model \model to mitigate the influence of the hidden confounder while solving the out-of-distribution generalization problem.  
	\item For efficient online detection, the two VAEs in CausalTAD calculate the posterior according to SD pairs and each road segment in the road network, which enables the update of anomaly score with the time complexity of $O(1)$. 
    % To boost the efficiency, we design a road-constrained VAE which directly learns the conditional probability $P(\bm{T}|\bm{C})$ and estimate the anomaly risk with partially observed trajectories. 
	\item Extensive experiments on two public trajectory datasets show that \model improves the anomaly detection performance of $2.1\% \sim 5.7\%$ and $10.6\% \sim 32.7\%$ on trajectories of observed and unobserved SD pairs, respectively. Moreover, \model is more efficient than the compared baselines.
\end{itemize}

%% file: sections/relatedwork.tex
\section{Related Work}
In this section, we first summarize the existing works on trajectory anomaly detection and out-of-distribution generalization, followed by a brief introduction to studies about causal inference.

\subsection{Trajectory Anomaly Detection}
The massive trajectories generated by various GPS devices have inspired the increasing focus on trajectory mining, \emph{e.g.}, trajectory clustering~\cite{LiCJPGH22,YangGMCW019,IJCNN17-YaoZZHB17}, trajectory similarity search~\cite{yao2019computing, li2018deep,DengZFSL022,YaoHDCHB22}.
Among these researches, trajectory anomaly detection is critical for ride-hailing platforms to detect taxi driving fraud and a series of methods have been proposed in recent years~\cite{ZhuJLLZ17,SuYB23,PAKDD18-ZhuYHLB18}.
We roughly divide existing trajectory anomaly detection methods into two categories: metric-based methods and learning-based methods.

\textbf{Metric-based Methods.}
Given a target trajectory, these methods define a set of reference trajectories with the same SD pair as the normal trajectories and calculate the distance metrics to normal trajectories as the anomaly scores.
There are methods that adopt Hausdorff distance~\cite{laxhammar2013online}, discrete Frechet distance~\cite{zhang2020continuous}, and other distance metrics to calculate the similarity between the target trajectory and the reference trajectories.
However, these methods always rely on thresholds that are sensitive to the distance metrics and the reference trajectories, thus it is difficult to select the appropriate thresholds in practice.
Furthermore, the efficiency of these methods is generally unsatisfactory due to the time-consuming distance metric calculation.

\textbf{Learning-based Methods.}
In recent years, there are lots of methods that utilize machine learning or deep learning to model the normal patterns from massive training trajectories.
For instance, DBTOD~\cite{wu2017fast} is a probabilistic model that adopts maximum
entropy inverse reinforcement learning to optimize the likelihood of driver behavior.
GM-VSAE~\cite{liu2020online} is a generative model that adopts VAE~\cite{kingma2013auto} to learn the trajectory pattern and utilize Gaussian mixture distribution to discover different types of normal routes.
DeepTEA~\cite{han2022deeptea} proposed a time-dependent method that encodes the varying traffic conditions at different times and locations into the latent variable.
The learning-based methods generally achieve better performance and efficiency than the metric-based ones, but they do not consider the bias caused by the road network.
Therefore, despite the success on in-distribution datasets, we find existing methods perform poorly on out-of-distribution datasets.

\subsection{Sub-Trajectory Anomaly Detection}
There is another series of methods that focus on a finer-grained task, \ie, sub-trajectory anomaly detection. While online trajectory anomaly detection requires timely identification of anomalies, sub-trajectory anomaly detection can provide better interpretability due to its fine-grained detection results. For example, ~\cite{lee2008trajectory} is a heuristic-based method for sub-trajectory anomaly detection. It partitions the current trajectory into segments and calculates the distance between each segment of the current trajectory and the segments in other trajectories. iBOAT~\cite{chen2013iboat} proposed an isolation-based method to isolate the anomalous sub-trajectories from the reference trajectories by maintaining an adaptive working window for the latest incoming GPS points. More recently, ~\cite{Zhang0LHYLCS23} proposes a novel reinforcement learning-based solution for sub-trajectory anomaly detection. It takes the proportion of occurrences of each road segment in trajectories with the same SD pair as the input feature. Then, it models these features with an LSTM and adopts a reinforcement learning model to determine whether each road segment is abnormal. Although these models provide finer-grained detection results, they all heavily rely on historical trajectories with the same SD pair, resulting in their inferior out-of-distribution generalization ability.

\subsection{Out-of-distribution Generalization}
There has been a lot of work to improve the model's ability to generalize out of distribution in different ways, such as domain generalization~\cite{LiYSH17,WangLLOQLCZY23,DuZ0S21}, stable learning~\cite{KuangCAXL18,Zhang0XZ0S21}, and invariant learning~\cite{CreagerJZ21}.
For instance, ~\cite{GaninUAGLLML16} proposed a domain adversarial neural network for learning representations that do not vary with the domain. ~\cite{peters2016causal} proposed invariant causal prediction, identifying the set of invariant variables in different environments through statistical testing, thereby improving the generalization ability. More recently, ~\cite{LiYSH18} divides the source domain into the meta-train domain and meta-test domain, and enhances the generalization ability of the model by applying meta learning. ~\cite{abs-1907-02893} proposed a representation-level invariance assumption, and based on this assumption, representations that can be generalized to different environments are learned by minimizing invariant risk. However, these methods require labels of the environments or domains, which are difficult to obtain in the trajectory anomaly detection task.

\subsection{Causal Inference}
While traditional machine learning methods only consider correlation among variables, they are always biased due to the confounders, resulting in poor stability and robustness.
Recently, lots of methods based on causal inference have emerged, which aim to identify the causal relations among variables instead of correlation.
They have been applied to various domains, such as computer vision~\cite{zhang2020causal, wang2020visual}, natural language processing~\cite{Yao0LX0Z19, keith2020text}, outlier detection~\cite{Davidson21}, online advertising~\cite{yuan2019improving, tan2022uncovering} and recommendation systems~\cite{wang2021deconfounded}, and achieved promising performance.
However, there is no existing work that studies the trajectory anomaly detection task from the causal perspective.
This paper finds for the first time that existing trajectory anomaly detection methods are essentially biased due to the confounder, resulting in poor generalization ability.
Due to the unobservable confounder and the efficiency requirements of online trajectory anomaly detection, existing causal inference techniques do not apply to our problem.
Therefore, we proposed \model to eliminate the confounding bias for stable and robust trajectory anomaly detection in an efficient way.

%% file: sections/preliminary.tex
\section{Preliminary}

\input{sections/preliminary/causalbasics}

\input{sections/preliminary/problem}

%% file: sections/preliminary/causalbasics.tex
\subsection{Causality Basics}
For the sake of clarity, we follow Pearl's framework~\cite{pearl2009causality} and describe the structural causal model briefly. Then, the \emph{do}-calculus is introduced to infer the causal effect through intervention.

\begin{figure}
    \centering
    \includegraphics[height=2.7cm]{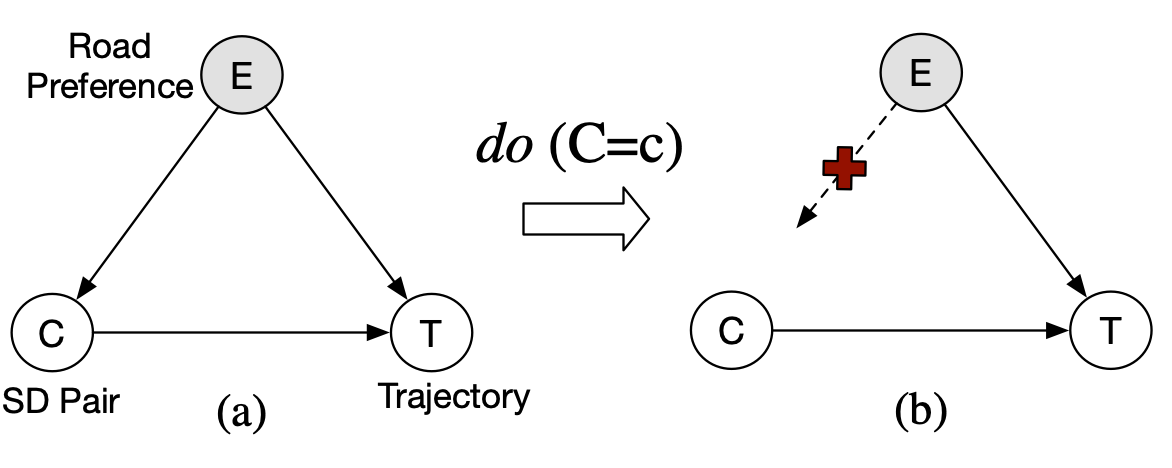}
    \vspace{-2ex}
    \caption{(a) An example of the causal graph. (b) The causal graph after an intervention.}
    \label{fig:causal-graph}
    \vspace{-2ex}
\end{figure}

\textbf{Causal Graph.}
A causal graph~\cite{glymour2016causal} describes the causal relations between variables as a directed acyclic graph $\mathcal{G} = \{\mathcal{V}, \mathcal{E}\}$.
$\mathcal{V}$ denotes the set of nodes, where each node corresponds to a variable.
$\mathcal{E}=\{(\bm{v}_i, \bm{v}_j)|\bm{v}_i,\bm{v}_j \in \mathcal{V}\}$ denotes the set of edges, where each edge $(\bm{v}_i, \bm{v}_j)$ points from the cause variable (or treatment variable) $\bm{v}_i$ to the result variable (or effect variable) $\bm{v}_j$.
For instance, Fig.~\ref{fig:causal-graph}(a) illustrates a typical causal graph, and each directed edge in the graph points from the treatment to the effect.

% \textbf{Structural Causal Model.} The Structural Causal Model (SCM) \cite{pearl2009causality} is defined as a graphical representation to formalize causal relationships. In SCM, causal relationships are expressed through deterministic, functional equations. 

% To be specific, given observed variables $\bm{X}=\{X_i\}_{i=1}^N$, an SCM $\mathcal{M}:=(\bm{F},\bm{X},\boldsymbol{\epsilon})$ consists of structural assignments $\bm{F}=\{f_i\}_{i=1}^N$,
% \begin{equation}
%     X_i = f_i(\bm{Pa}_i, \epsilon_i),
%     \nonumber
% \end{equation}
% where $\bm{Pa}_i$ is the set of parents of $X_i$, and each variable in $\bm{Pa}_i$ is a direct cause of $X_i$. $\boldsymbol{\epsilon}$ denotes noise variables.

% For every SCM, we can yield a \textit{causal graph} by adding one vertex for each $X_i$ and connecting an edge from each parent in $\bm{Pa}_i$ (the causes) to the child $X_i$ (the effect).
% Fig.~\ref{fig:causal-graph}(a) illustrates a typical causal graph, and each directed edge in the graph points from the cause to the effect.
% An essential attribute of causal graphs is \textit{Causal Markov Condition}~\cite{PearlV91}, which means the joint distribution of $\bm{X}$ can be factorized as $P(\bm{X}) = \prod_{i=1}^N P(X_i | \bm{Pa}_i)$.
% Based on this decomposition, we can derive some independence conditions between variables from the causal graph, \eg, $d$-separation criterion~\cite{glymour2008causal,pearl2009causality}.

\textbf{Confounding bias.}
A confounder is a variable that causes both the treatment and the effect. According to Causal Markov Condition~\cite{PearlV91}, the confounder can lead to a correlation between two variables, although there is no causal relationship between them, which is known as confounding bias. For instance, the temperature is a confounder for ice cream and crime rate. Higher temperature leads to both higher ice cream sales and a higher crime rate, resulting in the correlation between ice cream and crime rate. However, there is no causal relationship between ice cream and crime rate, thus the crime rate will not change with changes in ice cream sales. Therefore, if we train a generative model to predict the crime rate according to ice cream sales, it will fail when promotional activities change the distribution of ice cream sales. In other words, due to the existence of confounding bias, correlation cannot represent how a change in one variable affects another variable, which limits the out-of-distribution generalization ability of correlation-based prediction methods.

\textbf{Intervention.} While modeling correlation between variables may lead to confounding bias, we can adopt intervention to estimate causal effect instead of conditional probability. As shown in Fig.~\ref{fig:causal-graph}(a), the conditional probability $P(\bm{T}|\bm{C})$ models the correlation caused by both $\bm{C} \rightarrow \bm{T}$ and $\bm{C} \leftarrow \bm{E} \rightarrow \bm{T}$, but the correlation caused by $\bm{C} \leftarrow \bm{E} \rightarrow \bm{T}$ is the confounding bias and only $\bm{C} \rightarrow \bm{T}$ is the causal effect of $\bm{C}$ on $\bm{T}$. To eliminate the confounding bias and estimate the causal effect, the edge $\bm{E} \rightarrow \bm{C}$ can be cut off via intervention and \emph{do}-calculus~\cite{TianP02, ShpitserP08, pearl2009causality}, and we can estimate the causal effect as $P(\bm{T}|do(\bm{C}))$, as shown in Fig.~\ref{fig:causal-graph}(b). According to the \textit{backdoor criterion} \cite{glymour2016causal}, we can apply the adjust formula to $\bm{E}$ to calculate $P(\bm{T}|do(\bm{C}=\bm{c}))$ as follows:
\begin{equation}
\begin{aligned}
P(\bm{T}|do(\bm{C}=\bm{c})) = \sum_{\bm{e}} P(\bm{T}|\bm{C}=\bm{c}, \bm{E}=\bm{e})P(\bm{E}=\bm{e}).\label{eq:adjust}
\end{aligned}
\end{equation}

%% file: sections/preliminary/problem.tex
\subsection{Problem Definition}
In this section, we first present the definitions of some basic concepts and define the debiased online trajectory anomaly detection subsequently.

\define{1}{Trajectory}{A trajectory consists of an ordered sequence of points, starting at $S$ and ending with $D$. Each point in trajectory is a triplet $<x, y, time>$ representing the longitude, latitude, and timestamp.}

Note that trajectory anomalies considered in this work emerge in ride-hailing services. The sources and destinations, \ie SD pairs, are the prerequisites before the generation of trajectories. Moreover, ride-hailing orders mostly occur in the city indicating that precise road networks are available for anomaly detection. Thus, we assume all trajectories can be mapped into a completed road sequence.

\define{2}{Map-matched Trajectory}{Taking a trajectory and the related road network $\mathcal{G}$ as the input, the map-matched trajectory $\bm{t}$ is an ordered sequence of roads, \ie $\bm{t} = <t_1, t_2, \dots , t_n>$, where $t_i$ and $t_{i+1}$ are two adjacent road segments in the road network $\mathcal{G}$.}

We consider anomalies are trajectories that rarely occur and are different from other trajectories with the same SD pair constraint. In practice, online detection with ongoing trajectories is needed, which provides more chances for ride-hailing companies to prevent the anomaly from happening. Additionally, the generation of trajectory is confounded by a common cause, \ie the road network preference, severely affecting the generalization ability of current methods. As shown in Fig.~\ref{fig:causal-graph}(b), we employ $do$-calculus to eliminate the confounding bias. Thus, the problem in this paper can be defined as follows:

\define{3}{Debiased Online Trajectory Anomaly Detection}{Given an ongoing trajectory $\bm{t}$ with the SD pair $\bm{c}$, the task aims to infer whether the ongoing trajectory $\bm{t}$ would be an anomaly trajectory by computing $P(\bm{T}=\bm{t}|do(\bm{C}=\bm{c}))$ as the anomaly criterion.}

Different from existing online anomaly detection that aims to estimate the conditional probability $P(\bm{T}|\bm{C})$, 
our debiased online trajectory anomaly detection aims to estimate $P(\bm{T}|do(\bm{C}=\bm{c}))$, which denotes the causal relationship between $\bm{C}$ and $\bm{T}$. The conditional probability $P(\bm{T}|\bm{C})$ denotes the correlation between $\bm{C}$ and $\bm{T}$ and it is caused by both $\bm{C} \rightarrow \bm{T}$ and $\bm{C} \leftarrow \bm{E} \rightarrow \bm{T}$. However, the changes of $\bm{C}$ can not affect $\bm{T}$ through $\bm{C} \leftarrow \bm{E} \rightarrow \bm{T}$, resulting in the poor performance of existing methods when the distribution of $\bm{C}$ changes. Therefore, our debiased online trajectory anomaly detection adopts $do$-calculus to cut off $\bm{C} \leftarrow \bm{E} \rightarrow \bm{T}$ and estimates $P(\bm{T}|do(\bm{C}=\bm{c}))$ to improve the performance when the distribution of $\bm{C}$ changes.

It is worth noting that the debiased online trajectory anomaly detection is nontrivial. As shown in Eq.~\eqref{eq:adjust}, the calculation of $P(\bm{T}|do(\bm{C}))$ requires the distribution of $P(\bm{E})$, but $\bm{E}$ is unobserved since it denotes the mixture effects of many factors, such as the road level, speed limit, and even buildings on the road. Therefore, it is challenging to estimate $P(\bm{T}|do(\bm{C}))$ as the anomaly criterion.
Additionally, for online anomaly detection, the trajectory is being generated, thus $P(\bm{T}|do(\bm{C}))$ should be updated with time complexity of $O(1)$ when the ongoing trajectory reaches a new road segment.
It imposes a strict requirement for the efficiency of the model in debiasing.

%% file: sections/causalanalysis.tex
\section{Causal Analysis}

In this section, we first introduce the trajectory data generation process from the causal perspective. Then, we analyze why existing trajectory anomaly detection methods are essentially biased.
% and specify the challenges for eliminating the confounding bias.

We illustrate the structural causal model for trajectory generation in Fig.~\ref{fig:causal-graph}(a). As shown, the variable $\bm{C}$ refers to the SD pair of the trajectory $\bm{T}$, and the variable $\bm{E}$ refers to the urban layout and multiple factors of road segments in the city. Obviously, the trajectory should be generated according to the SD pair, \ie, $\bm{C} \rightarrow \bm{T}$. 
The causal relation $\bm{E} \rightarrow \bm{T}$ represents that trajectory generation would be influenced by the preference for roads. For instance, if there are multiple routes to reach the destination, drivers may prefer the main roads and avoid frequently congested road segments.
The causal edge $\bm{E} \rightarrow \bm{C}$ denotes the distribution of SD pairs is also affected by the road network preference.
For ride-hailing services, passengers tend to get in cars on parking-friendly paths and their destinations are usually some popular road segments, such as roads with malls and companies.
Moreover, the ride-hailing platforms also recommend some popular locations as the start and destination points, which further contribute to $\bm{E} \rightarrow \bm{C}$.

For detecting anomalies, existing methods calculate the conditional probability $P(\bm{T}|\bm{C})$ of trajectories as the anomaly criterion.
However, since $\bm{E}$ is a confounder that can cause the spurious correlation between $\bm{C}$ and $\bm{T}$, the detection results of existing methods are essentially affected by the confounding bias. Recall the example shown in Fig.~\ref{fig:causal_model}(b), due to the mall at $p_5$, most trajectories in the training dataset are destined for $p_5$. These trajectories prefer road $p_2-p_3$ over road $p_2-p_4$ and road $p_2-p_1$ according to $\bm{C}\rightarrow\bm{T}$ and $\bm{C} \leftarrow \bm{E} \rightarrow \bm{T}$. The conditional probability $P(\bm{T}|\bm{C})$ will model the preference for road $p_2-p_3$ and it works well for trajectory anomaly detection with destination $p_5$. However, during inference, there is a new destination $p_7$, which has never appeared in the training dataset. The SD pair $\bm{C}$ of this trajectory is outside the distribution of the training dataset, thus $\bm{C} \leftarrow \bm{E}$ does not hold.
Therefore, the correlation caused by $\bm{C} \leftarrow \bm{E} \rightarrow \bm{T}$ no longer exists for this trajectory, and $P(\bm{T}|\bm{C})$ will  overestimate the preference for road $p_2-p_3$ and underestimate the preference for road $p_2-p_4$.
As a result, existing methods that adopts $P(\bm{T}|\bm{C})$ as the anomaly criterion will assign a very low probability to the normal trajectory $p_2-p_4-p_6-p_7$ and predict it as an anomaly.
In conclusion, the conditional probability $P(\bm{T}|\bm{C})$ is essentially biased due to the confounder $\bm{E}$, resulting in the failure of existing methods in out-of-distribution generalization.

%% file: sections/methodology.tex
\section{Methodology}

In this section,  we first introduce the adjusting of the road preference bias, which decomposes the anomaly criterion into two parts, \ie likelihood and scaling factor. After that, we specify the two key modules of the proposed method \model which are used to estimate the likelihood and scaling factor respectively. Based on them, the debiased anomaly detection is provided to explain how to use \model to detect anomaly trajectories in an online manner. 
Finally, we provide an intuitive perspective on how \model eliminates the bias and we analyze the complexity and limitations of \model.

\input{sections/methodology/adjustment}

\begin{figure}
    \centering
    \includegraphics[height=5.4cm]{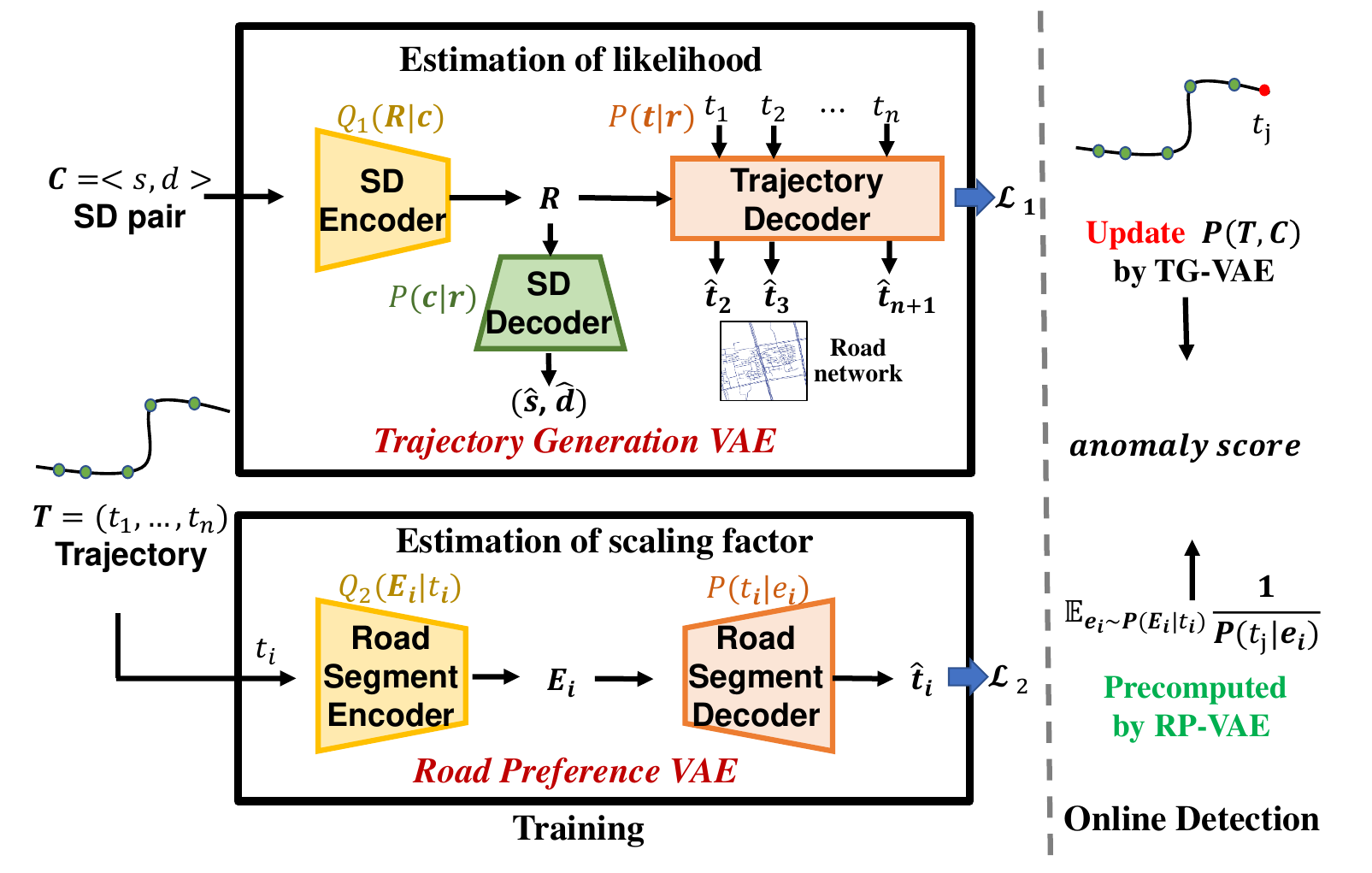}
    \vspace{-3ex}
    \caption{Overview of the \model. Upper-left: TG-VAE models the patterns of trajectories for each SD pair and estimates the likelihood for the tuple $(\bm{T}, \bm{C})$. Lower-left: RP-VAE factorizes the debiasing scaling factor to each road segment and estimates it via variational inference. Right: The debiased anomaly score can be calculated according to the likelihood estimated by TG-VAE and the scaling factor estimated by RP-VAE.}
    \label{fig:model}
    \vspace{-3ex}
\end{figure}

\input{sections/methodology/estimation1}
\input{sections/methodology/estimation2}

\subsection{Debiased Online Anomaly Detection}
According to the analysis above, the debiased anomaly score for a trajectory $\bm{t}=<t_1, \dots, t_n>$ with SD pair $\bm{c}$ can be calculated as:
\begin{equation}
\begin{aligned}
score &\overset{(1)}{=} -\log P(\bm{T}=\bm{t}|do(\bm{C}=\bm{c})) \\
&\overset{(2)}{=} -\log P(\bm{c}, \bm{t}) -\log \mathop{\mathbb{E}}\limits_{\bm{e} \sim P(\bm{E} | \bm{c}, \bm{t})} \frac{1}{P(\bm{c}|\bm{e})} \\
&\overset{(3)}{\approx} -\log P(\bm{c}, \bm{t}) - \lambda \sum_{i=1}^n \log \mathop{\mathbb{E}}_{\bm{e}_i \sim P(\bm{E}_i |t_i)} \frac{1}{P(t_i |\bm{e}_i)},
\end{aligned}
\label{equ:anomaly_score}
\end{equation}
where $(1)$ is because of our definition of debiased anomaly score, $(2)$ is because of Eq.~\eqref{eq:adjustment-for-segment}, $(3)$ is because of Eq.~\eqref{eq:factorize_scaling_factor} and Eq.~\eqref{eq:scaling_factor}, and $\lambda$ is a constant to balance the likelihood $P(\bm{c}, \bm{t})$ and the scaling factor.
The higher the score, the more likely the trajectory is to be abnormal.
To calculate the score, the \model can utilize TG-VAE and RP-VAE to estimate $P(\bm{c}, \bm{t})$ and $\mathbb{E}_{\bm{e}_i \sim P(\bm{E}_i |t_i)} 1/P(t_i |\bm{e}_i)$ respectively.

\model is efficient for ongoing trajectories to support online detection. Specifically, for TG-VAE, the trajectory encoder outputs the hidden variable $\bm{r}$ according to the SD pair $\bm{c}$, and the trajectory decoder predicts the next road segment based on the current hidden state $\bm{h}_j$ and current road segment $t_j$. Thus, it only has to generate the trajectory once, with the time complexity of $\mathcal{O}(1)$ for each input point. The RP-VAE is also efficient since it can calculate the scaling factor for each input point independently.

Furthermore, because the scaling factor has been factorized into each road segment independently, we can calculate and store the scaling factor for all road segments in the road network in advance to accelerate the inference. Thus we only have to go through TG-VAE to estimate $P(\bm{t}, \bm{c})$ during inference.

\input{sections/methodology/discussion}

%% file: sections/methodology/adjustment.tex
\subsection{Adjusting the road preference bias}
According to the causal graph shown in Fig.~\ref{fig:causal-graph}, we adopt the intervention $do(\bm{C}=\bm{c})$ to cut off $\bm{E} \rightarrow \bm{C}$ to eliminate the bias caused by $\bm{E}$.
Then we define the anomaly criterion of an observed trajectory $\bm{t}$ as $P(\bm{T}=\bm{t}|do(\bm{C}=\bm{c}))$.
According to the backdoor criterion, we can calculate $P(\bm{T}=\bm{t}|do(\bm{C}=\bm{c}))$ by applying the adjustment formula to $\bm{E}$:
\begin{equation}
\begin{aligned}
P(\bm{T}=\bm{t}|do(\bm{C}=\bm{c})) &= \sum_{\bm{e}} P(\bm{T}=\bm{t}|\bm{C}=\bm{c}, \bm{E}=\bm{e})P(\bm{e}) \\
&= P(\bm{c}, \bm{t}) \sum_{\bm{e}} \frac{P(\bm{e}|\bm{c}, \bm{t})}{P(\bm{c}|\bm{e})} \\
&= \underbrace{P(\bm{c}, \bm{t})}_{\textrm{ likelihood}} \underbrace{\mathop{\mathbb{E}}\limits_{\bm{e} \sim P(\bm{E} | \bm{c}, \bm{t})} {1}/{P(\bm{c}|\bm{e})}}_{\textrm{ scaling factor}}.
\label{eq:adjustment-for-segment}
\end{aligned}
\end{equation}
where $P(\bm{e})$ denotes $P(\bm{E}=\bm{e})$ for simplicity.

In Eq.~\eqref{eq:adjustment-for-segment}, there is a likelihood term $P(\bm{t}, \bm{c})$ and a scaling factor term. The likelihood $P(\bm{t}, \bm{c})$ is the likelihood of the observed trajectory, and the scaling factor term
% $\mathop{\mathbb{E}}\limits_{\bm{e} \sim P(\bm{E} | \bm{c}, \bm{t})} {1}/{P(\bm{c}|\bm{e})}$ 
indicating that debiased anomaly detection can be achieved by the appropriate scaling of the likelihood.
However, we must estimate $P(\bm{c}|\bm{E})$ and $P(\bm{E}|\bm{c}, \bm{t})$ to calculate the scaling factor, which is intractable due to the invisible $\bm{E}$. Therefore, in \model, we assume that the causal effect of $\bm{E}$ on $\bm{T}$ can be factorized into each road segment to deal with it. In the next subsection, we will specify how \model estimates the likelihood and scaling factor via its two modules respectively. For each module, we first give a theoretical derivation and then describe how to parameterize using neural networks.

%% file: sections/methodology/estimation1.tex
\subsection{Estimation of the likelihood}
\subsubsection{Theoretical Derivation}
We estimate $P(\bm{c}, \bm{t})$ via variational inference. Specifically, we estimate $P(\bm{c}, \bm{t})$ with a variational autoencoder, and the evidence lower bound (ELBO) of $\log P(\bm{c}, \bm{t})$ can be obtained through variational inference:
\begin{equation}
\begin{aligned}
\log P(&\bm{c}, \bm{t}) \geq \mathop{\mathbb{E}}\limits_{\bm{r} \sim Q_1(\bm{R}|\bm{c}, \bm{t})} \frac{P(\bm{c}, \bm{t}, \bm{r})}{Q_1(\bm{r}|\bm{c}, \bm{t})} \\
&=\mathop{\mathbb{E}}\limits_{\bm{r} \sim Q_1(\bm{R}|\bm{c}, \bm{t})} \log P(\bm{c}, \bm{t} | \bm{r}) - D_{KL}(Q_1(\bm{R}|\bm{c}, \bm{t})||P(\bm{R})),
% \mathop{\mathbb{E}}\limits_{\bm{r} \sim Q_1(\bm{R}|\bm{c}, \bm{t})}[ \log \frac{P(\bm{r}, \bm{c}, \bm{t})}{Q_1(\bm{r}|\bm{c}, \bm{t})}]\\
% &= \mathop{\mathbb{E}}\limits_{\bm{r} \sim Q_1(\bm{R}|\bm{c}, \bm{t})}[\log P(\bm{c}, \bm{t}|\bm{r}) + \log P(\bm{r}) -\log Q_1(\bm{r}|\bm{c}, \bm{t})] \\
\label{eq:eblo-route}
\end{aligned}
\end{equation}
where $\bm{R}$ is the latent variable, $P(\bm{R})$ is a pre-defined distribution that denotes the prior of the latent variable, $P(\bm{C}, \bm{T}|\bm{R})$ is the decoder, and $Q_1(\bm{R}|\bm{C}, \bm{T})$ is the encoder as well as the posterior.

For online anomaly detection, the trajectory $\bm{T}$ is always being generated during inference, thus $Q_1(\bm{R}|\bm{T}, \bm{C})$ needs to be updated frequently for each new input road segment, $\bm{r}$ should be resampled from $Q_1$ and the decoder $P(\bm{T}, \bm{C}|\bm{r})$ should restart decoding accordingly.
As a result, the time complexity of each new recorded point is $\mathcal{O}(n)$, where $n$ denotes the length of the ongoing trajectory.
To reduce time complexity, we adopt $Q_1(\bm{R}|\bm{C})$ rather than $Q_1(\bm{R}|\bm{T}, \bm{C})$ as the encoder during both training and inference.

We also apply mean-field approximation to factorize $P(\bm{c}, \bm{t}|\bm{r})$ as:
\begin{equation}
P(\bm{c}, \bm{t}|\bm{r}) = P(\bm{c}|\bm{r})P(\bm{t}|\bm{r}).
\nonumber
\end{equation}

Therefore, the ELBO in Eq.~\eqref{eq:eblo-route} can be written as:
\begin{equation}
\begin{aligned}
&\log P(\bm{t}, \bm{c}) \geq \\
&\mathop{\mathbb{E}}\limits_{\bm{r} \sim Q_1(\bm{R}|\bm{c})} [\log P(\bm{t} | \bm{r}) + \log P(\bm{c} | \bm{r})] - D_{KL}(Q_1(\bm{R}|\bm{c})||P(\bm{R})).
\label{eq:objective-route}
\end{aligned}
\end{equation}

As a result, there are two decoders $P(\bm{T}|\bm{R})$ and $P(\bm{C}|\bm{R})$ for trajectory and SD pair respectively. And we can take the ELBO in Eq.~\eqref{eq:objective-route} as the learning objective to jointly optimize these two decoders and the encoder $Q_1(\bm{R}|\bm{C})$.

\subsubsection{Trajectory Generation VAE}  According to the above analysis, we propose Trajectory Generation VAE, short for TG-VAE, to estimate $P(\bm{t}, \bm{c})$. In other words, given a trajectory $\bm{t} = <t_1, \dots, t_n>$ with SD pair $\bm{c}=<s, d>$, where $t_j$ is a road segment, the module should estimate the probability $P(\bm{t}, \bm{c})$. Different from traditional VAE, there are two improvements in TG-VAE.

\noindent\textbf{(1) Explicitly modeling of SD pair.} TG-VAE consists of an SD encoder $\Phi_{e}(\cdot)$, a trajectory decoder $\Phi_{t}(\cdot)$ and a SD decoder $\Phi_{c}(\cdot)$. The posterior of the latent variable $\bm{R}$ is calculated by $\Phi_{e}(\cdot)$ based on the SD pair $\bm{C}$ (instead of the trajectory $\bm{T}$).
In detail, we first map the discrete input to the continuous space:
\begin{equation}
\bm{s} = \mathbf{E}_c(s), \bm{d} = \mathbf{E}_c(d), \bm{x}^{(r)}_i = \mathbf{E}_r(t_i),
\nonumber
\end{equation}
where $\mathbf{E}_c$ and $\mathbf{E}_r$ are learnable embedding matrices, $\mathbf{E}_c(\cdot)$ and $\mathbf{E}_r(\cdot)$ denote embedding lookup, $s$ and $d$ is the source and destination respectively.
Then, the encoder infers the hidden representation $\bm{r}$ according to the source and destination:
\begin{equation}
\bm{\mu}_r, \bm{\sigma}_r = \Phi_e(\bm{s}, \bm{d}), \quad \bm{r} \sim \mathcal{N}(\bm{\mu}_r, \bm{\sigma}_r^2 \mathbf{I}).
\nonumber
\end{equation}
The $\Phi_e(\cdot)$ can be implemented as MLP or RNN.

After that, the SD decoder $\Phi_c(\cdot)$ predicts the source $\bm{\hat{s}}$ and the destination $\bm{\hat{d}}$ according to $\bm{r}$: 
\begin{equation}
\bm{\hat{s}}, \bm{\hat{d}} = \Phi_c (\bm{r}).
\nonumber
\end{equation}

In TG-VAE, we model the SD pair explicitly via the SD encoder and the SD decoder. Compared with calculating the posterior $\bm{R}$ with a trajectory encoder, the SD encoder has two advantages. Firstly, it avoids the continuous updating of the posterior while the trajectory is being generated in online detection. Secondly, it forces the model to utilize the SD pair to predict the trajectory. However, even with the SD encoder, the model may also predict trajectories based solely on popular transfer patterns, without considering the information of SD pairs. In this way, the posterior $\bm{R}$ becomes meaningless and collapses to the prior distribution. This posterior collapse problem results in a decrease in performance when the SD pair distribution changes. Therefore, in TG-VAE, the SD decoder predicts the SD pair according to the posterior to prevent the model from posterior collapse, thus improving the out-of-distribution generalization ability.

\noindent\textbf{(2) Road-constrained Prediction.} The trajectory decoder $\Phi_t$ employs RNN to predict the next road segment $\bm{\hat{t}}_j$ according to the hidden representation $\bm{r}$ and historical trajectory $\bm{x}^{(r)}_{<j}$:
\begin{equation}
\begin{aligned}
\bm{h}_{j+1} = &RNN(\bm{h}_{j}, \bm{x}^{(r)}_{j}), \quad 1\leq j < n, \\
&\bm{h}_0 = \bm{r}, \quad \bm{\hat{t}}^{(r)}_j = g(\bm{h}_j),
\nonumber
\end{aligned}
\end{equation}
where $g(\cdot)$ is the projection head and it only predicts $\bm{\hat{t}}^{(r)}_{j+1}$ from the neighbors of the current road segment $t_j$. In other words, we mask the road segments that are not neighbors of $\bm{t}_j$ when predicting $\hat{\bm{t}}_{j+1}$. Note that the confounding bias is caused by the imbalance of SD pairs in the training dataset, thus the popular SD pairs dominate the training of the road segment embeddings and the model ignores the information of other unpopular SD pairs. In our road-constrained prediction, the prediction of a popular road segment can only affect the road segments within its local area, which prevents trajectories of popular SD pairs from dominating the optimization of the whole road network. Therefore, the confounding bias has been alleviated.

According to Eq.~\eqref{eq:objective-route}, the loss of $\Phi(\cdot)$ can be defined as:
\begin{equation}
\begin{aligned}
\mathcal{L}_{1}(\bm{c}, \bm{t}) = &H(\bm{\hat{s}}, s) + H(\bm{\hat{d}}, d) + \sum_{i=1}^n H(\bm{\hat{t}}^{(r)}_i, t_i) \\
&+ D_{KL}(\mathcal{N}(\bm{\mu}_r, \bm{\sigma}_r^2 \mathbf{I}) || \mathcal{N}(\mathbf{0}, \mathbf{I})),
\nonumber
\end{aligned}
\end{equation}
where $H(\cdot)$ denotes the cross-entropy loss function. If well trained, we can estimate $Q_1(\bm{R}|\bm{c})$ with the SD encoder $\Phi_e(\cdot)$, estimate $P(\bm{c}|\bm{r})$ with the SD decoder $\Phi_c(\cdot)$, and estimate $P(\bm{t}|\bm{r})$ with the trajectory decoder $\Phi_t(\cdot)$. Then, we can approximate the likelihood term $P(\bm{c}, \bm{t})$ with the evidence lower bound in Eq.~\eqref{eq:objective-route}.

%% file: sections/methodology/estimation2.tex
\subsection{Estimation of the scaling factor} 

\subsubsection{Theoretical Derivation}
To calculate the debiased anomaly score $P(\bm{T}|do(\bm{C}))$, we have to estimate the scaling factor in Eq.~\eqref{eq:adjustment-for-segment}, which is intractable due to the invisible $\bm{E}$. Since $\bm{E} \rightarrow \bm{T}$ denotes how the preferences for each road segment affect the generation of $\bm{T}$, our basic idea is to estimate the scaling factor at the road level. We first convert $P(\bm{c}|\bm{e})$ to $P(\bm{t}|\bm{e})$ as:
\begin{equation}
\begin{aligned}
P(\bm{c}|\bm{e}) \overset{(1)}{=} \sum_{\bm{t}'} P(\bm{c}|\bm{t}', \bm{e}) P(\bm{t}'|\bm{e}) \overset{(2)}{=} \sum_{\bm{t}' \in \bm{T}(\bm{c})} P(\bm{t}'|\bm{e}),
\label{eq:p_t_e}
\end{aligned}
\end{equation}
where $(1)$ is because of the total probability theorem, $\bm{T}(\bm{c})$ denotes all trajectories whose SD pair is $\bm{c}$, $(2)$ is because if $\bm{t}' \in \bm{T}(\bm{c})$ then $P(\bm{c}|\bm{t}', \bm{e})=1$, otherwise $P(\bm{c}|\bm{t}', \bm{e})=0$. However, to calculate Eq.~\eqref{eq:p_t_e}, we have to traverse $\bm{T}(\bm{c})$, which is time-consuming and impractical. Fortunately, as shown in Eq.~\eqref{eq:adjustment-for-segment}, $\bm{e}$ is sampled from $P(\bm{E}|\bm{t}, \bm{c})$, thus $P(\bm{t}|\bm{e})$ will be much larger than $P(\bm{t}'|\bm{e})$ for $\bm{t}' \neq \bm{t}$. Therefore, we ignore terms in Eq.~\eqref{eq:p_t_e} where $\bm{t}' \neq \bm{t}$, and the scaling factor in Eq.~\eqref{eq:adjustment-for-segment} can be written as:
\begin{equation}
\begin{aligned}
\mathop{\mathbb{E}}\limits_{\bm{e} \sim P_1} \frac{1}{P(\bm{c}|\bm{e})} =  \mathop{\mathbb{E}}\limits_{\bm{e} \sim P_1} \frac{1}{\sum_{\bm{t}' \in T(\bm{c})} P(\bm{t}' | \bm{e})} \approx \mathop{\mathbb{E}}\limits_{\bm{e} \sim P_1} \frac{1}{P(\bm{t} | \bm{e})},
\label{eq:factorize_scaling_factor}
\end{aligned}
\end{equation}
where $P_1 = P(\bm{E} | \bm{c}, \bm{t})$. Since the $\bm{c}$ is useless for the rest calculation, we approximate $P(\bm{E}|\bm{c}, \bm{t})$ with $P(\bm{E}|\bm{t})$.
Recall that $\bm{E} \rightarrow \bm{T}$ denotes how the preferences for each road segment affect the generation of $\bm{T}$, we assume that the scaling factor can be factorized to each road segment. Therefore, we apply mean-field approximation to factorize $P(\bm{E}|\bm{T})$ and $P(\bm{T}|\bm{E})$, \ie, for $\bm{t}=<t_1, \dots, t_n>$:
\begin{equation}
\begin{aligned}
P(\bm{t}|\bm{e}) = \prod_{i=1}^n P(t_i |\bm{e}_i), \quad P(\bm{e}|\bm{t}) = \prod_{i=1}^n P(\bm{e}_i | t_i). \nonumber
\end{aligned}
\end{equation}
Then, the scaling factor term can be written as:
\begin{equation}
    \mathop\mathbb{E}_{\bm{e} \sim P(\bm{E}|\bm{c}, \bm{t}) }\frac{1}{P(\bm{c}|\bm{e})} \approx \prod_{i=1}^n \mathop\mathbb{E}_{\bm{e}_i \sim P(\bm{E}_i|t_i)} \frac{1}{P(t_i|\bm{e}_i)}.
    \label{eq:scaling_factor}
\end{equation}
We treat $\bm{e}_i$ as the latent variable and introduce a variational autoencoder to estimate the likelihood of each road segment $P(t_i)$. Then, we can approximate $P(\bm{E}_i|t_i)$ with the encoder in this VAE, and the decoder can estimate $P(\bm{t_i|\bm{e}_i})$. Specifically, the ELBO of $P(t_i)$ can be obtained through variant inference:
\begin{equation}
\log P(t_i) \geq \mathop{\mathbb{E}}\limits_{\bm{e}_i \sim Q_2(\bm{E}_i|t_i)} \log P(t_i|\bm{e}_i) - D_{KL}(Q_2(\bm{E}_i|t_i)||P(\bm{E}_i)),
\label{eq:objective-segment}
\end{equation}
where $P(\bm{E}_i)$ is a pre-defined prior, $Q_2(\bm{E}_i|t_i)$ is the posterior as well as the encoder, $P(t_i|\bm{E}_i)$ is the decoder. We can optimize them jointly by taking the ELBO as the learning objective.

\subsubsection{Road Preference VAE} Corresponding to the analysis above, there is a Road Preference VAE (RP-VAE) in \model. The RP-VAE $\Psi(\cdot)$ consists of a road segment encoder $\Psi_e(\cdot)$ and a road segment decoder $\Psi_d(\cdot)$.

Given a trajectory $\bm{t}=<t_1, \dots, t_n>$, we first map each road segment to the continuous space:
\begin{equation}
\bm{x}^{(s)}_i = \mathbf{E}_s(t_i),
\nonumber
\end{equation}
where $\mathbf{E}_s$ is a learnable embedding matrix.
Then, the encoder $\Psi_e(\cdot)$ takes each road segment as input and outputs the distribution of the latent variable $\bm{z}$:
\begin{equation}
\bm{\mu}^{(s)}_i, \bm{\sigma}^{(s)}_i = \Psi_e(\bm{x}^{(s)}_i), \quad \bm{e}_i \sim \mathcal{N}(\bm{\mu}^{(s)}_i, {\bm{\sigma}^{(s)}_i}^2 \mathbf{I}).
\nonumber
\end{equation}

After that, the decoder $\Psi_d(\cdot)$ predicts the road segment according to the latent variable:
\begin{equation}
\bm{\hat{t}}^{(s)}_i = \Psi_d(\bm{e}_i).
\nonumber
\end{equation}
Both the encoder $\Psi_e(\cdot)$ and the decoder $\Psi_d(\cdot)$ can be implemented via MLP.

According to Eq.~\eqref{eq:objective-segment}, the loss of $\Psi(\cdot)$ can be defined as:
\begin{equation}
\mathcal{L}_{2}(\bm{t}) = \sum_{i=1}^n \big [H(\bm{\hat{t}}^{(s)}_i, t_i) + D_{KL}(\mathcal{N}(\bm{\mu}^{(s)}_i, {\bm{\sigma}^{(s)}_i}^2 \mathbf{I}) || \mathcal{N}(\mathbf{0}, \mathbf{I}))\big],
\nonumber
\end{equation}
where $H(\cdot)$ denotes the cross-entropy loss function. If well trained, we can estimate $P(\bm{E}_i|t_i)$ with the road segment encoder $\Psi_e(\cdot)$ and estimate $P(t_i|\bm{e}_i)$ with the road segment decoder $\Psi_d(\cdot)$.

Based on the two modules, we adopt Adam~\cite{KingmaB14} as the optimizer to train \model.
Given the training dataset $\mathcal{D}=\{(\bm{c}_i, \bm{t}_i)|i=1,\dots, |\mathcal{D}|\}$, the parameters in TG-VAE and RP-VAE can be optimized jointly with the final loss:
\begin{equation}
\mathcal{L} = \sum_{i=1}^{|\mathcal{D}|}\mathcal{L}_1(\bm{c}_i, \bm{t}_i) + \mathcal{L}_2(\bm{t}_i).
\end{equation}

%% file: sections/methodology/discussion.tex
\subsection{Discussion}
\subsubsection{An Intuitive Perspective}
We have provided a detailed introduction to \model from a theoretical perspective, and here we explain why existing methods are biased and how \model can eliminate the bias from a more intuitive perspective for better understanding.

\textbf{Why are existing methods biased?} There are some popular locations in the city, such as malls, companies, and residential areas, which are always the destinations of trajectories. These trajectories occupy the majority of the training dataset, and dominate the training of the model. As a result, the model believes that the roads leading to these popular locations are better choices than others, which we refer to as popular roads. In other words, the model tends to underestimate the anomaly scores for trajectories passing through popular roads, and overestimate the anomaly scores for trajectories passing through other roads. 
If there is a distribution shift during inference, \ie, there are new destinations located on unpopular roads, the model tends to predict all these trajectories as anomalies.

\textbf{How does \model eliminate the bias?} \model eliminates the bias by weighting the anomaly scores of each road segment. Specifically, \model assigns higher weights for popular roads and lower weights for unpopular roads. Formally, we can rewrite the anomaly score in Eq.~\eqref{equ:anomaly_score} as:
\begin{equation}
    score \approx \sum_{i=1}^n -\log  \bigg(P(t_i|c,t_{<i})(\mathbb{E}_{e_i \sim P(\bm{E}_i|t_i)}\frac{1}{P(t_i|e_i)})^{\lambda}\bigg).
    \label{eq:intuition}
\end{equation}
In this equation, $-\log P(t_i|c,t_{<i})$ corresponds to the anomaly score of each road segment, and \model eliminates the bias by weighting the abnormal scores of different roads. Through this weighting, \model can compensate for the existing methods' underestimation of popular roads or overestimation of unpopular road segments.

\subsubsection{Complexity Analysis}
In this section, we discuss the complexity of \model in both training and online detection stages.  For a trajectory consisting of $n$ roads, we first analyze the time complexity of each module in model training.
For the TG-VAE, the time complexity of $\Phi_e$ and $\Phi_c$ is $\mathcal{O}(d)$, where $d$ is the hidden dimension.
The $\Phi_t$ only needs to generate the trajectory once, so the time complexity is $\mathcal{O}(dn)$.
For the RP-VAE, the time complexity of $\Psi_e$ and $\Psi_d$ is $\mathcal{O}(d)$, and each road segment is processed independently,
thus the time complexity for the trajectory is $\mathcal{O}(dn)$.
Therefore, the time complexity of the whole model is $\mathcal{O}(dn)$.

For online detection, \model takes the SD pair and an ongoing trajectory $\bm{t}$ as inputs, and calculates the anomaly score with Eq.~\eqref{equ:anomaly_score}. As described, the scaling factors estimated by RP-VAE could be calculated in advance. When a new road of $\bm{t}$ arrives, \model can reuse the previously computed probabilities and update the anomaly score in $\mathcal{O}(1)$ time. Thus, \model achieves high efficiency in online trajectory anomaly detection.

\subsubsection{Limitations and Future Work}
In this paper, we treat the variable $\bm{E}$ as static and assume that it does not change over time. However, a road may be congested during rush hours, but at other times it is clear. Therefore, $\bm{E}$ is dynamic and will change at different times of the day or on different days of the week.
While \model only factorizes the scaling factor to each road segment, we believe it's a better choice to factorize the scaling factor to each road segment and each time period. We remain it as future work.

%% file: sections/experiment.tex
\section{Experiment}
To validate the effectiveness of our method, we conduct extensive experiments to answer the following questions:
\begin{itemize}
    \item Can our method perform well not only on the trained SD pairs but also on unseen SD pairs?
    \item In the online detection setting, can our method achieve satisfactory performance and efficiency?
    \item How does each component of our model contribute to the performance?
    \item How do the hyperparameters affect the performance and how should we choose the best hyperparameters?
\end{itemize}
To answer the first question, experiments on in-distribution datasets and out-of-distribution datasets are conducted in Section~\ref{section:id} and Section~\ref{section:ood}, and we also evaluate the stability of \model on a mixture dataset in Section~\ref{section:stability}. To answer the second question, we conduct online evaluation and efficiency evaluation in Section~\ref{section:online} and Section~\ref{section:eff}. To answer the third and the last question, Section~\ref{section:eff} and Section~\ref{section:param} give the ablation study and analysis of hyperparameters. All the code and data have been released\footnote{\href{https://github.com/LwbXc/CausalTAD}{https://github.com/LwbXc/CausalTAD}}.

\input{tables/table1}

\input{sections/experiment/setup}

\input{tables/table2}

\input{sections/experiment/in-distribution}

\input{sections/experiment/out-of-distribution}

\begin{figure}
    \centering
    \includegraphics[height=3.5cm]{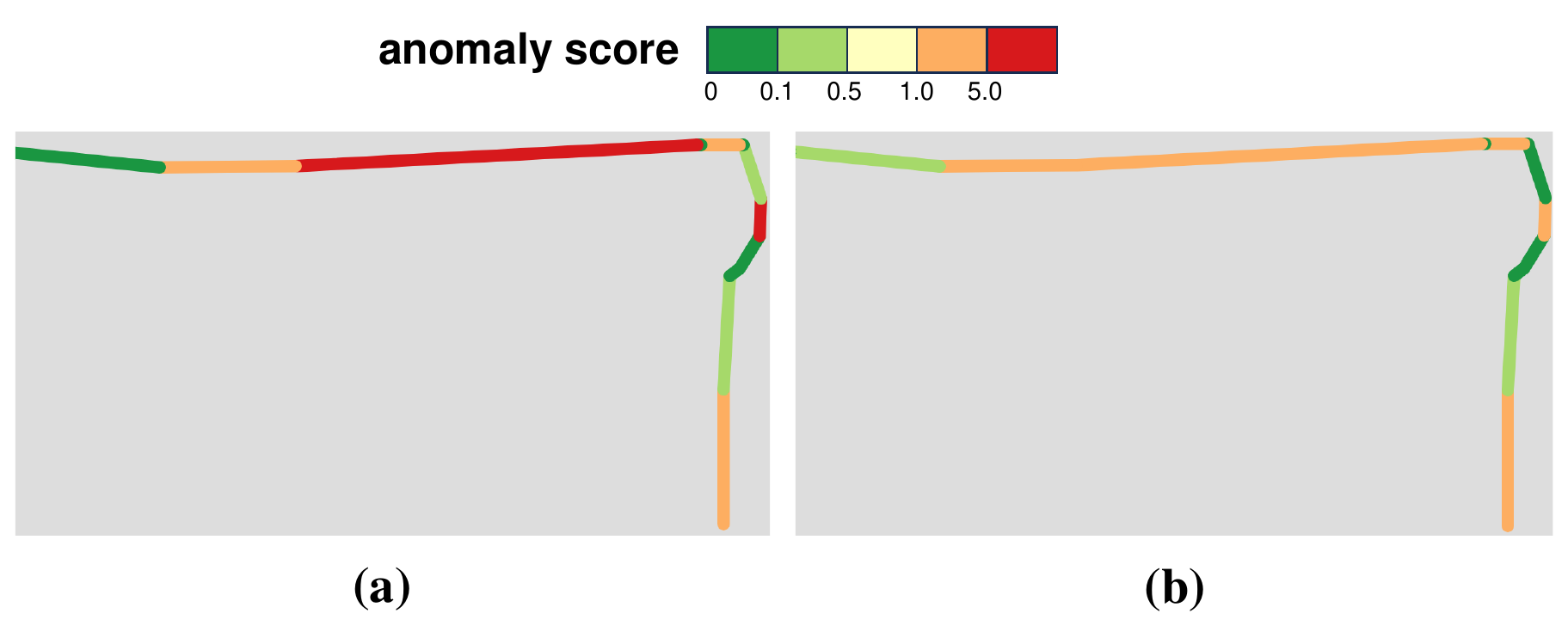}
    \vspace{-2ex}
    \caption{The anomaly scores of a normal trajectory with an unseen SD pair estimated by (a) VSAE and (b) CausalTAD.}
    \label{fig:casestudy}
    \vspace{-4ex}
\end{figure}

\begin{figure}
    \centering
    \includegraphics[height=3.5cm]{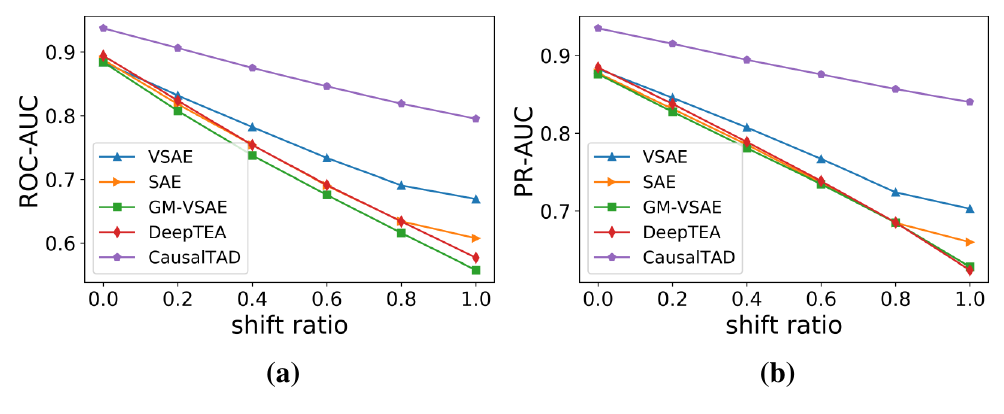}
    \vspace{-2ex}\caption{Performance under different ratios of distribution shift. (a) The ROC-AUC on Detour dataset of Xi'an. (b) The PR-AUC on Detour dataset of Xi'an.}
    \label{fig:mixture}
    \vspace{-3ex}
\end{figure}

\input{sections/experiment/stability}

\input{sections/experiment/online}

\begin{figure}
    \centering
    \includegraphics[height=3.5cm]{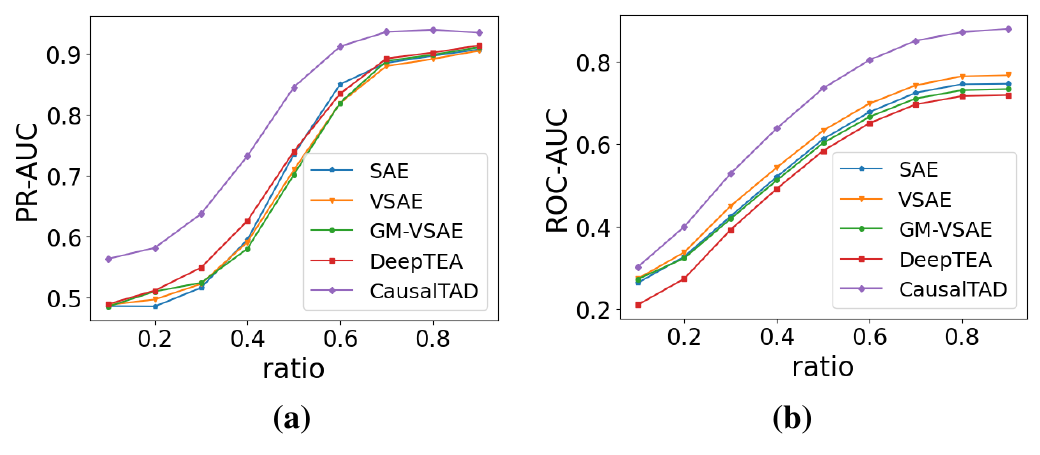}
    \vspace{-2ex}
    \caption{Performance under different observed ratios. (a) ID \& Switch datasets of Xi'an. (b) OOD \& Switch datasets of Chengdu.}
    \label{fig:online-evaluation}
    \vspace{-2ex}
\end{figure}

\begin{figure}
    \centering
    \includegraphics[height=3.5cm]{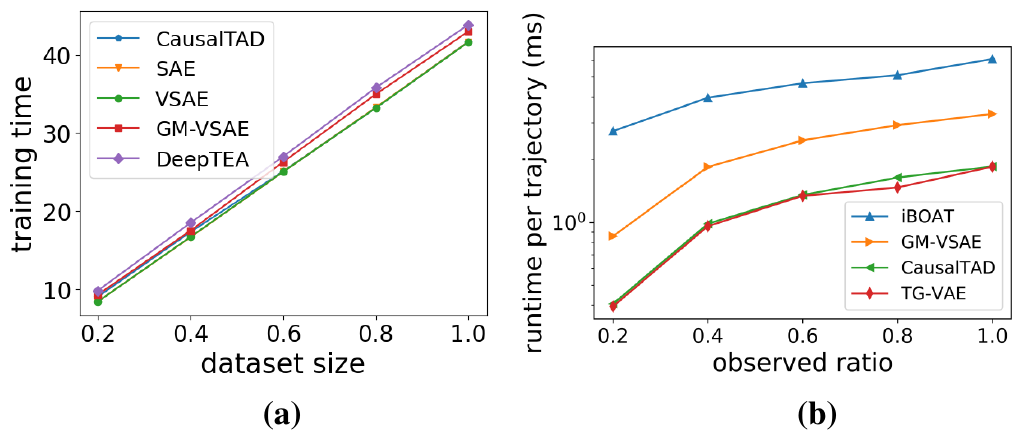}
    \vspace{-2ex}
    \caption{(a) Comparison of training scalability. (b) The average runtime per trajectory during inference under different observed ratios.}
    \label{fig:efficiency-evaluation}
    \vspace{-3ex}
\end{figure}

\input{tables/table3}
\begin{figure*}
    \centering
    \includegraphics[height=3.3cm]{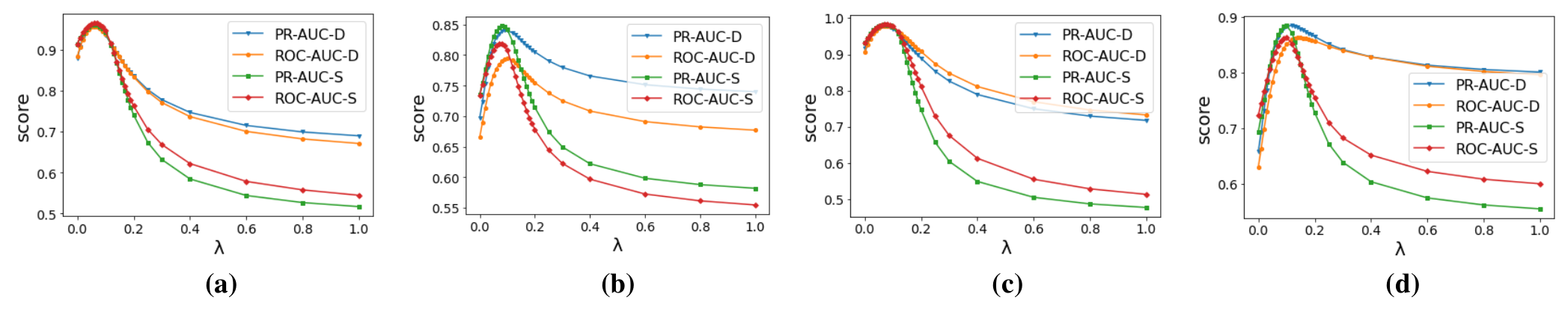}
    \vspace{-1ex}
    \caption{Performance of \model under different values of $\lambda$. "-D" and "-S" denote Detour and Switch. (a)In-distribution datasets of Xi'an. (b)Out-of-distribution datasets of Xi'an. (c)In-distribution datasets of Chengdu. (d)Out-of-distribution datasets of Chengdu.}
    \label{fig:parameter-analysis}
    \vspace{-3ex}
\end{figure*}

\input{sections/experiment/efficiency}

\input{sections/experiment/ablation}

\input{sections/experiment/parameter}

%% file: tables/table1.tex
\setlength{\abovecaptionskip}{0.cm}
\setlength{\belowcaptionskip}{-0.2cm}
\renewcommand{\arraystretch}{1.6}
\begin{table*}[ht]  
    \centering  
    \fontsize{9}{7.5}\selectfont
    \setlength\tabcolsep{3pt}
    \caption{The performance of \emph{ROC-AUC}, \emph{PR-AUC} on the in-distribution datasets (bold: best; underline: runner-up.) }
    \label{tb:id-evaluation} 	
    \begin{tabular}{l|c|c|c|c|c|c|c|c}  
    \bottomrule
     & \multicolumn{4}{c|}{Xi'an}  &  \multicolumn{4}{c}{Chengdu} \cr
    \cline{2-9}
     & \multicolumn{2}{c|}{ID \& Detour} & \multicolumn{2}{c|}{ID \& Switch} & \multicolumn{2}{c|}{ID \& Detour} & \multicolumn{2}{c}{ID \& Switch} \cr
     \cline{2-9}
     & ROC-AUC & PR-AUC & ROC-AUC & PR-AUC
     & ROC-AUC & PR-AUC & ROC-AUC & PR-AUC \cr
     \hline
     iBOAT
     & 0.4660 & 0.5119 & 0.7318 & 0.7084 
     & 0.4717 & 0.4947 & 0.4948 & 0.4965\cr
     VSAE
     & 0.8837 & 0.8827 & 0.9114 & 0.9129
     & 0.9130 & 0.9219 & 0.9358 & 0.9342 \cr
     SAE
     & 0.8875 & 0.8773 & 0.9174 & 0.9097 
     & 0.9279 & 0.9368 & \underline{0.9451} & \underline{0.9461}\cr
     $\beta$-VAE & 0.8693 & 0.8690 & 0.9072 & 0.9098 & 0.8836 & 0.8774 & 0.9120 & 0.8996 \cr
     FactorVAE & 0.8463 & 0.8520 & 0.8946 & 0.9000 & 0.8988 & 0.9076 & 0.9206 & 0.9149 \cr
     GM-VSAE
     & 0.8837 & 0.8770 & 0.9147 & 0.9153
     & \underline{0.9292} & \underline{0.9412} & 0.9444 & 0.9451 \cr
     DeepTEA
     & \underline{0.8939} & \underline{0.8845} & \underline{0.9238} & \underline{0.9231}
     & 0.9228 & 0.9346 & 0.9437 & 0.9423 \cr
     \hline
     \model
     & \textbf{0.9371} & \textbf{0.9351} & \textbf{0.9463} & \textbf{0.9424}
     & \textbf{0.9745} & \textbf{0.9716} & \textbf{0.9788} & \textbf{0.9757} \cr
    \hline
    Improvement
    & 4.8\% & 5.7\% & 2.4\% & 2.1\%
    & 4.9\% & 3.2\% & 3.6\% & 3.1\% \cr
    \toprule  
    \end{tabular}
    \vspace{-4ex}
\end{table*}

%% file: sections/experiment/setup.tex
\subsection{Experimental Setup}
\subsubsection{Dataset}
We evaluate our model on two public real-world datasets from DiDi Inc, which contain the taxi trajectories of Xi'an and Chengdu respectively. All trajectories have been matched into the road network, and we filter out trajectories shorter than 30. For each city, we sample 100 SD pairs with more than 100 trajectories as candidate pairs. We randomly select half of the trajectories of the 100 candidate pairs as the \textbf{training dataset}, and the other half as the \textbf{ID test dataset}. To validate the effectiveness of our model on SD pairs never seen before, we randomly sample trajectories from the whole dataset as the \textbf{OOD test dataset.} Finally, there are about 10,000 trajectories in each dataset of Xi'an and about 20,000 trajectories in each dataset of Chengdu.

\subsubsection{Ground Truth}
Since no labeled dataset is available for trajectory anomaly detection, there are two choices for ground truth, \ie, manual labeling and anomaly generation. However, due to the existence of elevated highways, turning restrictions on roads, and dynamic changes in traffic conditions, it is difficult to obtain reliable labels via manual labeling. Additionally, manual labeling is very time-consuming, which limits  the size of the dataset used for evaluation. Therefore, following previous work~\cite{han2022deeptea, liu2020online}, we generate anomalous trajectories for evaluation. While the anomaly-generation strategies in previous work are designed for grid trajectories, it does not work for road segment trajectories under the constraint of the road network. Here we introduce two anomaly-generation strategies on the road network. In detail, for a trajectory $\bm{t}=\{t_1, \dots, t_n\}$ where $t_i$ denotes a road segment:
\begin{itemize}[leftmargin=4mm]
    \item \textbf{Detour.}
    We create a detour trajectory by temporarily deleting a road segment from the road network. Specifically, we choose three indexes $1\leq i < k < j \leq n$ and temporarily delete $t_k$ from the road network, and apply Dijkstra algorithm~\cite{dijkstra2022note} to obtain the shortest path from $t_i$ to $t_j$. Then we get a detour trajectory by replacing the sub-trajectory $\{t_i, t_{i+1}, \dots, t_j\}$ with this path. We can find a detour trajectory with appropriate detour distance by traversing all possible $i$, $k$, and $j$.
    \item \textbf{Switch.}
    We create a switch trajectory by switching from one route to another different route. Specifically, given a trajectory $\bm{t}$, we first find out the trajectories of the same SD pair in the whole dataset and then sample a trajectory $\bm{t}'$ from those with a low similarity score. For trajectory $\bm{t}'$, the similarity score is defined as $|\bm{t}' \cap \bm{t}|/|\bm{t}' \cup \bm{t}|$, where $\cap$ and $\cup$ denote intersection and union of the road segment set. Then we create an anomaly trajectory by switching from $\bm{t}$ to $\bm{t'}$.
\end{itemize}
Based on the above strategies, we generate two anomalous datasets, namely \textbf{detour test dataset} and \textbf{switch test dataset}. Four combinations of these anomalous datasets and normal datasets (\emph{i.e.}, ID test dataset, and OOD test dataset) are used as the final test datasets.

\subsubsection{Evaluation Metrics}
Following previous works~\cite{chen2013iboat, liu2020online, han2022deeptea}, we adopt Precision-Recall AUC (PR-AUC) and Receiver Operating Characteristic AUC (ROC-AUC) as the metrics, which are commonly used in anomaly detection tasks.

\subsubsection{Baseline}
To prove the superiority of our method, we compare it with several trajectory anomaly detection methods. We also include two disentanglement methods to test whether disentanglement can improve the generalization ability.
\begin{itemize}
    \item \textbf{iBOAT}~\cite{chen2013iboat} is a metric-based method, which detects anomalies by comparing the test trajectory against historical trajectories with the same SD pair.
    \item \textbf{VSAE} adopts the framework of basic VAE~\cite{kingma2013auto} but implements the encoder and decoder as RNN.
    \item \textbf{SAE}~\cite{malhotra2016lstm} is a traditional Seq2Seq model which reconstructs the input trajectories and takes the reconstruction error as the anomaly score.
    \item \textbf{$\beta$-VAE}~\cite{HigginsMPBGBML17} introduces a hyperparameter to VAE to balance latent channel capacity and independence constraints, which might improve its generalization ability.
    \item \textbf{FactorVAE}~\cite{KimM18} is a method that disentangles by encouraging the distribution of representations to be factorial and hence independent across the dimensions.
    \item \textbf{GM-VSAE}~\cite{liu2020online} detects anomalous trajectories via a generative model. It adopts a sequential variant autoencoder to capture complex information enclosed in trajectories and Gaussian mixture distribution to discover different types of normal routes.
    \item \textbf{DeepTEA}~\cite{han2022deeptea} is a time-dependent method that adopts the framework of a generative model and captures the dynamic traffic conditions.
\end{itemize}

\subsubsection{Experiment Parameters}
All the experiments are conducted on a single NVIDIA RTX2080Ti GPU. The hidden dimension $d$ is set to 128. For the learning-based methods, we train each method for 200 epochs with an initial learning rate $0.01$ and report the results of the model performing best on the validation dataset.
After the grid search, we set the constant $\lambda$ in our method to 0.1. This experiment is detailed in Section \ref{section:param}.

%% file: tables/table2.tex
\setlength{\abovecaptionskip}{0.cm}
\setlength{\belowcaptionskip}{-0.2cm}
\renewcommand{\arraystretch}{1.6}
\begin{table*} 
    \centering  
    \fontsize{9}{7.5}\selectfont
    \setlength\tabcolsep{3pt}
    \caption{The performance of \emph{ROC-AUC}, \emph{PR-AUC} on the out-of-distribution datasets (bold: best; underline: runner-up.) }
    \label{tb:ood-evaluation} 	
    \begin{tabular}{l|c|c|c|c|c|c|c|c}  
    \bottomrule
     & \multicolumn{4}{c|}{Xi'an}  &  \multicolumn{4}{c}{Chengdu} \cr
    \cline{2-9}
     & \multicolumn{2}{c|}{OOD \& Detour} & \multicolumn{2}{c|}{OOD \& Switch} & \multicolumn{2}{c|}{OOD \& Detour} & \multicolumn{2}{c}{OOD \& Switch} \cr
     \cline{2-9}
     & ROC-AUC & PR-AUC & ROC-AUC & PR-AUC 
     & ROC-AUC & PR-AUC & ROC-AUC & PR-AUC \cr
     \hline
     iBOAT
     & 0.1408 & 0.3960 & 0.3793 & 0.4689
     & 0.3569 & 0.4318 & 0.3617 & 0.4093 \cr
     VSAE
     & \underline{0.6692} & \underline{0.7029} & \underline{0.7302} & \underline{0.7354}
     & \underline{0.6417} & 0.6673 & \underline{0.7344} & \underline{0.7029} \cr
     SAE
     & 0.6075 & 0.6598 & 0.6896 & 0.6998
     & 0.6350 & \underline{0.6676} & 0.7176 & 0.6948 \cr
     $\beta$-VAE & 0.5945 & 0.6623 & 0.6818 & 0.7129 & 0.5506 & 0.5880 & 0.6576 & 0.6237 \cr
     FactorVAE & 0.5647 & 0.6379 & 0.6664 & 0.6999 & 0.5731 & 0.6186 & 0.6740 & 0.6443 \cr
     GM-VSAE
     & 0.5574 & 0.6279 & 0.6497 & 0.6697
     & 0.6164 & 0.6639 & 0.7020 & 0.6842 \cr
     DeepTEA
     & 0.5771 & 0.6239 & 0.6684 & 0.6699
     & 0.5698 & 0.6082 & 0.6842 & 0.6520 \cr
     \hline
     \model
     & \textbf{0.7950} & \textbf{0.8401} & \textbf{0.8077} & \textbf{0.8421}
     & \textbf{0.8514} & \textbf{0.8839} & \textbf{0.8636} & \textbf{0.8844} \cr
    \hline
    Improvement
    & 18.8\% & 19.5\% & 10.6\% & 14.5\%
    & 32.7\% & 32.4\% & 17.6\% & 25.8\% \cr
    \toprule
    \end{tabular}
    \vspace{-4ex}
\end{table*}

%% file: sections/experiment/in-distribution.tex
\subsection{In-distribution Evaluation\label{section:id}}
To demonstrate that our model can capture the trajectory generative probability for SD pairs in the training dataset and effectively identify anomalous samples, we conduct experiments on two dataset combinations: \textbf{ID test dataset \& detour test dataset} and \textbf{ID test dataset \& switch test dataset}. For each dataset combination, the number of anomalous trajectories is similar to normal trajectories. The distribution of SD pairs of the selected test dataset is the same as that of the training dataset, which is exactly the setup in the previous work. The results are presented in Table~\ref{tb:id-evaluation}.

As the results show, our \model achieves the best performances on trajectories of both Xi'an and Chengdu, with an improvement $2.4\%\sim4.9\%$ on ROC-AUC and $2.1\%\sim5.7\%$ on PR-AUC. This indicates that segment bias exists not only in datasets out of the distribution but also in in-distribution datasets. For the baseline models, the metric-based iBOAT performs poorly on most datasets. For example, both metrics on the datasets of Chengdu are near 0.5, which is not better than random guesses. This is because the metric-based method fails to cope with the context information in the real world. $\beta$-VAE and FactorVAE perform worse than other learning-based methods because they sacrifice some of their modeling capacity to achieve independence across the dimensions. The other four baselines, GM-VSAE, DeepTEA, SAE, and VSAE, show comparable performance on all in-distribution datasets. This mainly owing to the Seq2Seq model they adopted, which is powerful in learning normal patterns from trajectories. We find that GM-VSAE and DeepTEA perform better than other baselines in most cases, which indicates both dynamic traffic conditions and encoding different types of normal routes via Gaussian mixture distribution are beneficial to anomaly detection.

%% file: sections/experiment/out-of-distribution.tex
\subsection{Out-of-distribution Evaluation\label{section:ood}}
While previous works evaluate their methods on in-distribution datasets, there will always be new SD pairs in practical applications, which require robustness to out-of-distribution trajectories. To verify the effectiveness of our model when applied to datasets with new SD pairs, we conduct experiments on two dataset combinations: \textbf{OOD test dataset \& detour test dataset} and \textbf{OOD test dataset \& switch test dataset}. For each dataset combination, the number of anomalous trajectories is similar to normal trajectories. The metric-based method needs reference trajectories with the same SD pair during inference, which are always unavailable in the OOD dataset. Thus, for a new SD pair $\bm{c}$, we take the trajectories whose SD pair is closest to $\bm{c}$ as reference trajectories. The results are shown in Table~\ref{tb:ood-evaluation}.

Since there are lots of new SD pairs whose trajectory patterns can not be learned directly from the training datasets, the performance of all baselines drops significantly. For example, the ROC-AUC and PR-AUC of iBOAT are below 0.5, which denotes it can not distinguish between normal trajectories with new SD pairs and anomalous samples. The methods based on the Seq2Seq model also perform poorly, with a reduction of $20\% \sim 40\%$ compared with the in-distribution evaluation. We also find that SAE and VSAE perform better than GM-VSAE and Deep-TEA. It indicates the dynamic traffic conditions and the Gaussian mixture distribution of different route types are beneficial to fit the in-distribution trajectory pattern but sacrifice the ability of out-of-distribution generalization. As for the $\beta$-VAE and FactorVAE, their performance is inferior to VSAE in all settings, which indicates that disentanglement cannot solve the problem of out-of-distribution generalization in trajectory anomaly detection.

Despite the performance drop of all baselines, our method still achieves satisfactory performance and outperforms all baselines by a large margin. For example, the ROC-AUC and PR-AUC on Chengdu are between 0.85 to 0.89, with an improvement of $17.6\% \sim 32.7\%$ compared with the best baseline. The results demonstrate that our model can implicitly learn the trajectory pattern of new SD pairs from the training dataset, thus distinguishing between the new SD pairs and the anomalous samples.

To illustrate how CausalTAD improves its performance on OOD test datasets, we visualize the anomaly scores of a normal trajectory with an unseen SD pair in Fig.~\ref{fig:casestudy}. Note that we have centralized the scaling factor part of the anomaly scores. Due to the unseen destination, the trajectory passes through some unpopular road segments that rarely appear in the training dataset. SAE assigns anomaly scores larger than 5 to them and predicts this normal trajectory as anomalous. As shown in Eq.~\eqref{eq:intuition}, CausalTAD assigns a higher $\mathbb{E}_{e_i \sim P(\bm{E}_i|t_i)}\frac{1}{P(t_i|e_i)}$ for these unpopular road segments, which compensate for the overestimation of the anomaly scores. As a result, CausalTAD predict it as normal.

In conclusion, the experimental results prove the superiority of our model in both in-distribution fitting and out-of-distribution generalization.

%% file: sections/experiment/stability.tex
\subsection{Stability Evaluation}\label{section:stability}

Unlike the OOD test dataset, the trajectories in the real world only partially undergo distribution shifts. Therefore, we mix the ID test dataset and the OOD test dataset to evaluate the stability of \model at different distribution shift levels. Specifically, we introduce a parameter $\alpha$ as the shift ratio and we mix the ID test dataset and the OOD test dataset in a ratio of $1-\alpha$ to $\alpha$. We conduct experiments on the Detour dataset of Xi'an with different $\alpha$ and the results are shown in Fig.~\ref{fig:mixture}.

We observe that the performance of all methods decreases linearly as the shift ratio $\alpha$ increases, which is in line with our intuition. All baselines show similar performance when the shift ratio is zero, but the performance of VSAE decreases more slowly than other baselines. When the shift ratio increases to 0.2, the performance of VSAE has already surpassed other baselines and this gap becomes larger as the shift ratio increases. \model has the slowest performance degradation and consistently outperforms all other baselines, which demonstrates the stability of \model.

%% file: sections/experiment/online.tex
\subsection{Online Evaluation\label{section:online}}

Online detection is critical in real-world applications since it supports the platform with more time and a higher probability to prevent anomalies from escalating into severe incidents.
Thus we conduct a set of online experiments to investigate the effectiveness of our proposed \model. Specifically, the experiments study how the performance changes when different ratios of trajectories are observed. The observed ratio measures the percentage of observations that the evaluated methods can utilize, and online evaluations are equal to offline evaluations when the observed ratio is set to $1.0$. From the previous experiments, we regard learning-based baselines (SAE, VASE, GM-VASE, DeepTEA) as competitors of \model, therefore, we compare \model with them in the online environment. A part of our results is shown in Fig.~\ref{fig:online-evaluation}, which covers the performance on ID \& Switch datasets of Xi'an and OOD \& Switch datasets of Chengdu. We find that:
% {\color{blue} The complementary results for different types of anomalies and cities can be found in the appendix.}
 
(1) The performances of all evaluated approaches go up as the observed ratio grows. This is because the models tend to conduct more accurate results based on more complete observations. The curve is relatively flat at the beginning, while the slope increases in the middle stage. This caters to the factor that anomalies most occur in the middle of the trajectories in our datasets.

(2) The four learning-based baselines show comparable performance and \model outperforms them consistently at all values of observed ratios.
And the performance gap between \model and learning-based baselines is widening especially on OOD datasets, as shown in Fig.~\ref{fig:online-evaluation}(b). 

(3) Moreover, the performance of \model reaches a decent degree when the observed ratio equals $0.6$, \ie, the PR-AUC is $0.9123$ in Fig.~\ref{fig:online-evaluation}(a) and the ROC-AUC is $0.8030$ in Fig.~\ref{fig:online-evaluation}(b).
Whereas, the learning-based baselines derive comparable performances when the observed ratio is $[0.8,1.0]$.

In conclusion, \model provides trustworthy anomaly detection earlier on all datasets, which verifies its superiority for online anomaly detection compared with other methods.

%% file: tables/table3.tex
\setlength{\abovecaptionskip}{0.cm}
\setlength{\belowcaptionskip}{-0.2cm}
\renewcommand{\arraystretch}{1.6}
\begin{table*} 
    \centering  
    \fontsize{9}{7.5}\selectfont
    \setlength\tabcolsep{3pt}
    \caption{Ablation-study results on
     the two parts: TG-VAE and RP-VAE.}
    \label{tb:ablation-study} 	
    \begin{tabular}{l|c|c|c|c|c|c|c|c|c}  
    \bottomrule 
     \multicolumn{2}{c|}{} & \multicolumn{4}{c|}{Xi'an}  &  \multicolumn{4}{c}{Chengdu} \cr
    \cline{3-10}
     \multicolumn{2}{c|}{} & \multicolumn{2}{c|}{ID}  &  \multicolumn{2}{c|}{OOD} & \multicolumn{2}{c|}{ID} & \multicolumn{2}{c}{OOD}\cr
    \cline{3-10}
     \multicolumn{2}{c|}{} & Detour & Switch & Detour & Switch & Detour & Switch & Detour & Switch \cr
     \hline
     \multirow{2}{*}{CausalTAD} & PR-AUC & 0.9351 & 0.9424 & 0.8401 & 0.8421 & 0.9716 & 0.9757 & 0.8839 & 0.8846\cr
     & ROC-AUC & 0.9371 & 0.9463 & 0.7950 & 0.8077 & 0.9745 & 0.9788 & 0.8514 & 0.8636\cr
     \hline
     \multirow{2}{*}{TG-VAE} & PR-AUC & 0.8797 & 0.9139 & 0.6975 & 0.7356 & 0.9197 & 0.9309 & 0.6581 & 0.6940\cr
     & ROC-AUC & 0.8834 & 0.9141 & 0.6657 & 0.7342 & 0.9067 & 0.9315 & 0.6301 & 0.7236 \cr
     \hline
     \multirow{2}{*}{RP-VAE} & PR-AUC & 0.6531 & 0.4833 & 0.7230 & 0.5604 & 0.6696 & 0.4441 & 0.7819 & 0.5309 \cr
     & ROC-AUC & 0.6280 & 0.4986 & 0.6562 & 0.5293 & 0.6723 & 0.4574 & 0.7716 & 0.5701 \cr
     \toprule
    \end{tabular}
    \vspace{-3ex}
\end{table*}

%% file: sections/experiment/efficiency.tex
\subsection{Efficiency Evaluation\label{section:eff}}
\textbf{Training scalability.} We first investigate the training scalability of \model, SAE, VSAE, GM-VSAE, GM-VSAE, and DeepTEA.
We vary the size of the training dataset of Xi'an from $20\%$ to $100\%$ and train all models with a single NVIDIA RTX2080Ti GPU.
Fig.~\ref{fig:efficiency-evaluation}(a) shows the results.  
From the figure, we can see that all methods scale linearly w.r.t. the size of training data. 
Therefore, they're promising to handle massive trajectory data. 

\textbf{Inference evaluation.} To process massive daily trajectories online, the model should be efficient during inference. Fig.~\ref{fig:efficiency-evaluation}(b) presents the average runtime of detecting a trajectory under different observed ratios. 
The metric-based method, iBOAT, is much slower than other learning-based methods. The reason is that comparing the current trajectory and historical trajectories results in its high computational complexity and it cannot be accelerated with GPUs.
Among the learning-based methods, \model and TG-VAE are consistently faster than GM-VSAE under all observed ratios since their time complexity is $\mathcal{O}(d)$ for each input point during inference.
The runtime of \model is very close to TG-VAE, which indicates the time cost of debiasing is very small and can be ignored.

%% file: sections/experiment/ablation.tex
\subsection{Ablation Study\label{setion:ablation}}
To answer how each component of the proposed method contributes to the performance, we compare \model with two ablation methods, \ie, TG-VAE and RP-VAE. As described in our methodology, the TG-VAE aims to learn the trajectory pattern for each SD pair and RP-VAE estimate the scaling factor for each road segment.
The results are shown in Table~\ref{tb:ablation-study}.
We find that the PR-AUC and ROC-AUC of TG-VAE and RP-VAE are inferior to \model, which proves that both components help improve the performance. 
We can observe the performance of the TG-VAE is much better than the RP-VAE in most settings, which indicates that the trajectory pattern for each SD pair is more informative compared to road segments.
This result is consistent with our cognition.

%% file: sections/experiment/parameter.tex
\subsection{Parameter Analysis\label{section:param}}
The hyperparameter $\lambda$ is critical to balance the likelihood of the trajectory and the scaling factor. To answer how performance changes under different values of $\lambda$ and how to select an appropriate value, we conduct experiments on all dataset combinations of both cities. The results are shown in Fig.~\ref{fig:parameter-analysis}. We observed that:

(1) When $\lambda$ is set to 0, the performance is satisfactory on the in-distribution datasets but poor on the out-of-distribution datasets, which is similar to the baselines based on the Seq2Seq model.
\model would degrade to VSAE when $\lambda$ is 0.

(2) As $\lambda$ increases, the metrics on all dataset combinations increase first. It indicates the idea of factorizing the scaling factor to each road segment is successful. But when $\lambda$ is too large, \emph{e.g.}, $\lambda=1$, the performance drops significantly, which shows that an appropriate $\lambda$ is crucial for our methods.

(3) For all datasets combinations of both cities, when $\lambda$ is near 0.1, the metrics always reach the maximum value. As shown in Eq.~\eqref{eq:factorize_scaling_factor}, we overlooked some terms in the denominator and the scaling factor is actually overestimated. Therefore we have to choose a small $\lambda$ to compensate for this overestimation. While setting $\lambda$ to 0.1 is the best choice for our test datasets, we recommend conducting the grid search on the validation dataset to determine the best value of $\lambda$ for other datasets.

%% file: sections/conclusion.tex
% \vspace{-2ex}
\section{Conclusion}
In this paper, we are the first to analyze the trajectory generation process under the casual perspective and define a new task, \ie debiased online trajectory anomaly detection, to eliminate the confounding bias introduced by the road preference. Consequently, a causal implicit generative model \model is proposed to solve the new task. \model employs \emph{do}-calculus to break the back door of confounder and estimate the unbiased anomaly criterion $P(\bm{T}|do(\bm{C} = \bm{c}))$. \model consists of two key generative models, \ie trajectory generation VAE (TG-VAE) and road preference VAE (RP-VAE), which are designed to estimate the likelihood and scaling factor, respectively. Based on them, \model can achieve superior performance on both effectiveness and efficiency. We have conducted extensive experiments on two public trajectory datasets. The results show that \model obtains significant performance improvements of $2.1\% \sim 5.7\%$ and $10.6\% \sim 32.7\%$ on trajectories of observed and unobserved SD pairs, compared to the state-of-the-art baselines. 